# Believing in BERT:
# Using expressive communication to enhance trust and counteract operational error in physical Human-Robot Interaction *

Adriana Hamacher[a], Nadia Bianchi-Berthouze[b], Anthony G. Pipe[c] and Kerstin Eder[d]

*Abstract*— Strategies are necessary to mitigate the impact of unexpected behavior in collaborative robotics, and research to develop solutions is lacking. Our aim here was to explore the benefits of an affective interaction, as opposed to a more efficient, less error prone but non-communicative one. The experiment took the form of an omelet-making task, with a wide range of participants interacting directly with BERT2, a humanoid robot assistant. Having significant implications for design, results suggest that efficiency is not the most important aspect of performance for users; a personable, expressive robot was found to be preferable over a more efficient one, despite a considerable trade off in time taken to perform the task. Our findings also suggest that a robot exhibiting human-like characteristics may make users reluctant to 'hurt its feelings'; they may even lie in order to avoid this.

## I. INTRODUCTION

The new arena of collaborative robotics is sorely in need of strategies to deal with the challenges that arise where robots and humans work in close proximity. The human-populated world contains an infinite number of unknowns. Thus, while collaborative robotics is still in its nascent phase, designing for the initial mistakes, misunderstandings and failures likely to arise between human and robot is crucial. How does a robot recover a user's trust after an error? How effective is an attempt to rectify the situation, or an apology, in mitigating dissatisfaction caused by unpredictable behavior? To what extent can a machine displaying human-like attributes soften displeasure? These are the central questions this paper seeks to begin to address.

Typically, evaluation has tended to emphasize the efficiency of collaboration under varying conditions, as outlined in [27]. This (simulated) study suggests that efficiency is not the most important aspect of performance for users and seeks to demonstrate that embodied emotional expressiveness improves the integration of human-robot activity. Taking this one step further, we explored whether expressiveness - verbal and non-verbal - could mitigate any dissatisfaction caused by erroneous or unexpected behavior and positively affect the participants' experience.

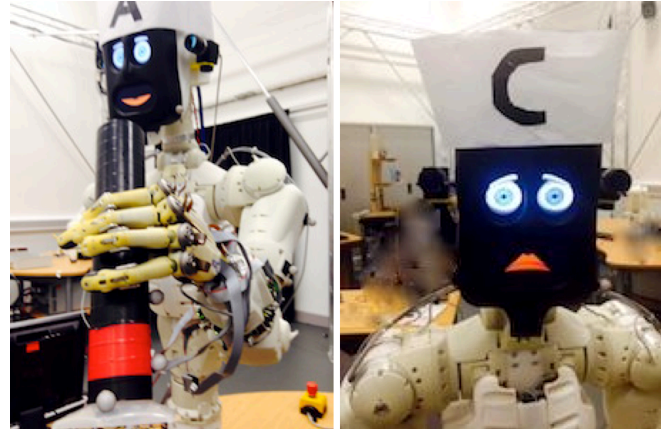

*Figure 1: The BERT2 platform* with neutral expression (left) and BERT C's facial expression on egg drop (right).

Our study took the form of a real-life task, with participants spanning a range of ages and levels of experience (with robots), working directly with a humanoid robot assistant (Fig. 1) in an omelet-making task. A total of 15 of the 21 participants preferred the communicative, personable robot over a more efficient, less error prone, but non-communicative one. Satisfaction was significantly increased in the communicative condition and participants were particularly responsive to this robot's apparent awareness of its error and expression of regret. For the majority, personable, transparent behavior appeared to negate the fact that the interaction took 50 per cent longer than in the non-communicative conditions.

Our results suggest that users are likely to prefer an expressive and personable robot, even if it is less efficient and more error prone, than a non-communicative one. Our study furthermore offers fresh insight into how mistakes made by a robot affect its trustworthiness and acceptance in human-robot collaboration. It suggests that a robot effectively demonstrating apparent emotions, such as regret and enthusiasm, and awareness of its error, influences the user experience in such a way that dissatisfaction with its erroneous behavior is significantly tempered, if not forgiven, with a corresponding effect on trust. In fact, human-like characteristics may make users reluctant to hurt a robot's 'feelings' and they may even lie in order to avoid this.

## II. LITERATURE REVIEW

As robots are increasingly developed for use in social settings, acceptance, persuasiveness and likability are key and these are factors strongly linked to trust [32]. Trust thus emerges as a central force in this study and requires definition. We also look at the role of efficiency and the impact of speech and embodied expressiveness.

* This work was supported in part by the EPSRC grants EP/K006320/1 and EP/K006223/1, as part of the project "Trustworthy Robotic Assistants."

[a] A. Hamacher is with the UCL Interaction Centre, University College London, UK. She is also a freelance journalist (+44 7808 719739; email: adriana.hamacher.11@ucl.ac.uk, adahamacher@gmail.com).

[b] N. Bianchi-Berthouze is with the UCL Interaction Centre, University College London, UK (email: n.berthouze@ucl.ac.uk).

[c] A.G. Pipe is with the Bristol Robotics Laboratory, Bristol, UK (email: Tony.Pipe@brl.ac.uk).

[d] K. Eder is with the Department of Computer Science, University of Bristol, UK (email: kerstin.eder@bristol.ac.uk) and leads the Verification and Validation for Safety in Robots research theme at the Bristol Robotics Laboratory, Bristol, UK.

*A. Trust and Performance*

Trust in automation is thought to be largely based on the extent to which a machine is perceived to properly perform its function [25]. This implies that machine errors strongly affect trust. However, trust is a complicated and multidimensional construct. It develops in a combination of three interplaying processes: analytic, analog, and affective [20]. This affective notion of trust aligns with psychological theory that, rather than competence, it is trust that is the primary quality people seek when evaluating somebody they encounter for the first time [7]. Competence, they say, is evaluated only after trust is established.

Human Robot Interaction (HRI) studies (e.g. [4]), have found that errors occasionally performed by a humanoid robot can actually increase its perceived human-likeness and likability. However, we lack large-scale, long-term data on the effects of unexpected behavior in HRI, despite the fact that many experiments go wrong in more ways than they go right.

*B. Developing Trust*

One approach to gaining trust, in complex systems, is good 'etiquette', described [24] as the *"largely unwritten codes that define roles and acceptable or unacceptable behaviors or interaction moves of each participant in a common 'social' setting."* Good etiquette has been found to significantly enhance diagnostic performance, regardless of reliability [28]. Its effects were powerful enough to overcome low reliability with a corresponding effect on trust. While not wishing to detract from the importance of reliability, our study seeks to explore this idea. It asks whether good etiquette, manifest in an expressive, communicative interface, that appears aware of its error, may be a way to compensate for initial mistakes that may be made by collaborative robots.

*C. Methods of Communicating*

Donald Norman calls lack of transparency 'silent automation' [26]. For users, transparency and control may even be more important than increased autonomy [16], so a system should clearly communicate its intention in a timely manner [8]. However, effectively achieving this can be a minefield in robotics (see, for example, [18] where inappropriate terminology caused even more confusion). While many studies urge caution in respect of verbal communication in HRI [10], developers of robots for the elderly have found that other methods of communication, such as gesture control or on-screen cues, can be equally problematic for those with infirmities [35].

Much social robotics research is in agreement that the main requirements of a complex social interaction include communication, particularly for the elderly [14], the recognition and expression of emotions, and some rudimentary form of personality [10]. These features are widely thought to increase the believability of artificial agents and enhance engagement. Prior studies suggest that users prefer speech [30, 15] and react favorably to non-verbal communication [5], including facial expression [33]. This influenced our decision to rely on these as a means of achieving a more satisfactory interaction.

There is evidence too that people are appreciative of robots that apologize or offer compensation if they have made a mistake [21], although this has not previously been explored in a real life setting with a humanoid robot as we seek to do here.

III. METHOD[1]

'Believing in BERT' was an experiment undertaken to explore viable strategies a humanoid robot might employ to counteract the effect of unsatisfactory task performance. Participants were invited to select a robot kitchen assistant from three potential job candidates. The same robot, BERT2, was used, but differentiated as candidate A, B or C (letters were judged most neutral, although it is recognized that even this type of labelling may lead to bias). Each participant was exposed to the robot acting in all three conditions:

BERT A: Non-communicative, most efficient.

BERT B: Non-communicative, makes a 'mistake' and attempts to rectify it.

BERT C: Communicative, expressive, also makes a 'mistake' and attempts to rectify it.

We investigated participants' subjectively self-reported and objectively measured behavioral data.

*A. Hypotheses*

Based on our literature review, we developed the following Hypotheses:

1. An unforeseen occurrence will cause a robot to appear less trustworthy than a more reliable one, even if attempts are made to mitigate the mistake (i.e. BERT A will be more popular than BERT B).

2. Increased transparency and feedback, manifest in communication and facial expression, can significantly mitigate dissatisfaction in the event of an unforeseen occurrence (i.e. BERT C will be more popular than BERT B).

3. Given the choice between enhanced efficiency and reliability or a personable, communicative interface, most people will choose the later (i.e. BERT C will be chosen over BERT A).

*B. Experimental Design*

A within-subjects design was adopted and the order of the job candidates - A, B and C - was counterbalanced and varied in a chi-square. The independent variables were the efficiency of the robot and its communicative ability. By comparing the effect of our experimental conditions, we hoped to shed light on what inspires confidence in an agent. The length of interactions with BERT's A and B was fixed and approximately the same, but BERT C was programmed to communicate with participants, which meant that the length of the interaction increased and was dependent on whether speech was recognized and repetition needed. In order to minimize confounds, C was programmed to recognize only "yes" and "no" answers and participants were informed of this limitation.

---

[1] Underlying data are openly available from the University of Bristol's Research Data Repository, https://data.bris.ac.uk/, under the DOI: 10.5523/bris.1xkj9m7z4vi6l137gihezyd9o3. Not all data used in this study can be made available due to ethical concerns.

Approval was first obtained from the University of the West of England Ethics Review Committee. The experiment then took place at the Bristol Robotics Laboratory (BRL) in the UK. The platform used was BERT2, an upper-body humanoid robot, with seven degrees-of-freedom (DOF) for each arm and hand [23]. Its suitability to interact with humans safely and naturally [19] motivated its selection. Voice recognition used the CSLU Toolkit and Rapid Application Development (RAD) with TCL scripting language. RAD uses the Festival speech synthesis system and recognition is based on Sphinx-II [23]. BERT2's face is a hybrid, combining an expressive digital interface with a static human visage-like structure and is capable of multiple variations [2]. The two we chose are visible in Fig. 1, with the "default" expression in all the robots on the left and the facial expression, presented only in condition C, after dropping the egg, on the right.

### C. Experimental Procedure

After signing a consent form, participants were individually invited into a curtained-off room, with BERT2 situated at the center, adjacent to two tables. They first completed a pre-experiment questionnaire. The objective was to gather data, such as their age and their experience of working with robots, and a 5-point Likert scale was used. They were then instructed to stand next to BERT2 in a mock cooking scenario. An (A4, 498-word) information sheet was provided [11] and they were told that a robot would be handing them ingredients to make an omelet. They were asked to evaluate different versions of the same robot and answer any questions it may ask, using the microphone provided. With the non-speaking robots, it was up to the participants to decide how to collaborate. A cap was placed on the robot's head to indicate which of the assistants was being tested - A, B or C.

On the table furthest from the participant were placed four (polystyrene) eggs, and, on the other, in front of him or her, a bowl and whisk. A video camera was sited in front of the scene and the researcher was seated just in front of it. A prominent safety button was within participants' reach and, also for reasons of safety, the interaction was slower than a human-human handover task would be.

The robot was pre-programmed and acted autonomously in each condition. The non-communicative candidate, labeled A, performed the most efficiently, never dropping an egg. B was also mute and, in addition, dropped one of the eggs. It attempted to rectify this by trying the handover again, using a different method to present the egg palm up (as opposed to dropping it into the participant's hand, the default position). After successfully handing over three eggs, all the robots passed the participants a container of salt and then invited them to whisk the eggs.

BERT C was the only candidate able to talk. On each occasion the system verified whether participants were ready to receive the egg *("are you ready for the egg"* or *"are you ready for the salt?"*), with action dependent on reply *("yes"* or *"no";* in the latter case the system would repeat the question). BERT C also dropped an egg, but appeared conscious of its mistake and apologized. It then attempted to rectify the error and forewarned participants that it would try another method of handover. At the end of the task, it asked the participants whether it did well and whether it got the job.

The rationale behind these conditions was to gauge the effect of increased transparency and human-like attributes in a situation characterized by uncertainty and vulnerability.

Following their interaction with the robot, participants completed a post-experiment questionnaire based on the NASA TLX. A 5-point Likert scale was used, with two additional measures for satisfaction and trust phrased as follows: "On a scale of 1-5… how satisfied were you with BERT A/B/C?" and "On a scale of 1-5… how much would you trust BERT A/B/C?". Participants were then invited to choose one of the robots for the job. Finally, they were interviewed, with questions focused on how effective and/or engaging the kitchen assistants were, and then debriefed. The duration of the experiment was approximately 50 minutes.

### D. Analytical Methods

In line with other studies [32] that examine the effect of error in humanoid robotics, we measured trust based on self-reported, quantitative questionnaire data, as well as on behavioral data that assesses trust based on the participants' willingness to cooperate.

*1) Self reported data:*

SPSS for Mac was used for all statistical analyses of Likert scale questionnaire data. Means and standard deviations were derived for the pre-experimental data used to develop some of our dependent variables.

The post-experiment questionnaire scale data were not normally distributed; non-parametric Friedman tests were used to investigate the effect of independent variables on the Hypotheses. We used Wilcoxon matched pairs tests to compare the effects of the conditions, where the Friedman tests demonstrated significance, and applied a Bonferroni correction to the significant findings in order to counter the likelihood of chance results.

*2) Behavioral and interview data:*

Initial video analysis of behavioral data followed the approaches developed by [17] and [13]. Analysis took in elements such as length of pauses, loud or soft speech, quickening and slowing of pace, gaze, orientation, gesture and postural movement. Participants' willingness to cooperate with the robot [31] and the amount of attention it required [1] were further considerations.

The ELAN[2] platform was used for behavioral coding of verbal and non-verbal actions observed during video analysis and semi-structured interviews.

Achieving inter-rater correlation proved challenging (agreement between two observers who coded 10% of the behavioral data was low, at 0.250) so, rather than a systematic analysis of all the video, we used the annotation performed to seek evidence supporting or questioning the self-reported data in which our findings are grounded. This approach is systematically used in ethnography to better understand qualitative accounts. Indeed, it allowed us to discover patterns and relationships, to contextualize our observations and to complement, illustrate and provide support for the self-reported data. Categories were developed inductively, taking into account what participants had told us, frequency of observation, valence and critical junctions in the interaction (Table 1).

---

[2] http://tla.mpi.nl/tools/tla-tools/elan/

TABLE I. BEHAVIORAL CATEGORIES

| Behavior | Physical example | Verbal example |
|---|---|---|
| Emotional reactions and explicit references to feelings. | Uncertainty, surprise, annoyance. | "The expression did affect me." |
| Responses that rationalise participants' reactions to a response. | Not following the robot's suggestion to whisk the eggs ("it wasn't real enough.") | "It was reassuring that B and C presented a solution." |
| Behaviour indicating confusion, embarrassment or indecisiveness. | Looking away. | "When BERT tried the open palm method of handing the egg over, that was confusing." |
| Statements and body language resulting from the robot's unreliability. | Leaning back. | "I was disappointed that the speed was so slow and the behavior so error prone." |
| Reactions or responses attributing intelligence to the robot or praising it. | Smiling at or mimicking the robot's behaviour. | "B seemed to be able to learn to be better. His delivery of the next egg was more clever." |
| Empathetic behaviour towards the robot. | Helping the robot by attempting to rescue an egg. | "Thank you!" |
| Negative behaviour, indicative of impatience or dissatisfaction. | Hand(s) on hip, scratching chin. | "[It was] better with speech." |

*E. Participants*

A total of 23 participants, 12 men and 11 women, were recruited from the local area and from BRL. Care was taken to achieve a wide range of ages, from 22-72, and a mixture of naive users, those with some experience of robots and robotics and other students. Data from two subjects was ultimately discarded. In both cases the robot was malfunctioning to the point where the subjects could not complete the tasks.

## IV. RESULTS

The results are divided into three sections: the pre-experiment questionnaire, which provides a profile of the participants, the post-experiment questionnaire, showing the impact on our Hypotheses, and interview data integrated with behavioral analysis, to validate the self-reported data.

*A. Pre-Experiment Questionnaire*

Our participants were spread across a wide range of age groups (M = 41.14, SD = 73.62). Over half (52.28%) were employed, with the rest evenly spread between students, retired and unemployed.

Ten participants (six females) had had no exposure to robots either at work or home. Of these, four (two females) picked the most efficient robot, BERT A, and six (four females) chose the communicative one, BERT C, as the candidate they would choose to work with again. So neither their experience with robots nor their gender appeared to influence their ultimate choice.

When asked to rate their level of dissatisfaction when things went wrong on a scale of 1-5 (with 5 representing the highest level of dissatisfaction), the most popular rating was 4 (M = 3.66, SD = 18.98), indicating low tolerance levels.

Our pre-experiment questionnaire data also serves to highlight how important users' preconceptions are. Participants were given a number of options for things that they considered BERT could and couldn't do and their answers were far from accurate. 95.24% credited the robot with recognizing speech, although only 90.48% thought it could vocalize. 38.10% believed that they would be able to have a conversation with the robot. Two participants said they thought BERT would be able to recognize mood and three believed it could juggle objects.

In the post-experiment questionnaire, two of the respondents who said they believed BERT would be able to juggle scored maximum on feeling insecure, stressed and annoyed with all the BERTs, although C scored slightly lower here. Their satisfaction scores for all three robots were very low and they said they were unlikely to want to use any of them again.

*B. Post-Experiment Questionnaire*

BERT C was the preferred candidate overall and 15 respondents said they would give this robot the job. Their reasons were largely based on the communicative abilities possessed by C, making feedback possible. For example: *"The vocal interaction with BERT C stopped me wondering what was happening next. It also let me know when he realized that he had dropped the egg. With the non-vocal machines there is a nervousness about when I should be holding out my hand, etc."*

The six candidates that chose BERT A all referred to the robot's robustness as the reason for their selection. For example: *"Bert A made the fewest mistakes."* One participant who chose A said they would prefer C if it were a work - as opposed to home related - scenario and, of the participants who chose BERT C overall, two said they would reconsider and choose BERT A or B, if the task was work related.

Turning to the Likert scale data, Friedman test results (with Kendall's W as a measure of effect size) showed that, although mental and physical demand, performance and effort did not vary significantly among the three conditions, the effect of the type of robot used was, however, seen for the other measures. In particular, satisfaction ($\chi2(2) = 18.353$, $p = 0.001$; Kendall's W = 0.437), temporal demand ($\chi2(2) = 14.000$, $p = 0.001$; Kendall's W = 0.350), trust ($\chi2(2) = 10.226$, $p = 0.006$; Kendall's W = 0.243) and frustration ($\chi2(2) = 8.442$, $p = 0.015$; Kendall's W = 0.201) varied significantly in the three conditions (Mean ranks in Fig. 2).

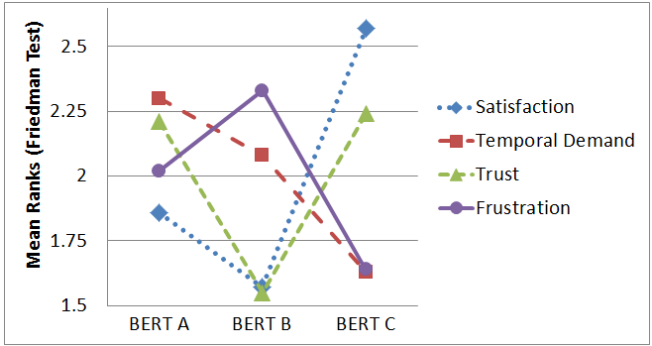

Figure 2: Mean ranks for Satisfaction, Temporal Demand, Trust and Frustration as resulted from the Friedman test.

Post hoc Wilcoxon Signed Rank tests using the Bonferroni correction revealed a statistically significant reduction in temporal demand, with BERT C, (M = 1.20, SD = 0.69) when compared with BERT A (M = 1.75, SD = 1.06); Z = -2.810, p = 0.005. Levels of perceived satisfaction also differed significantly between B (M = 2, SD = 0.94) and C (M = 2.76, SD = 1.30); Z = -2.799, p = 0.005). While interesting, the A/C results did not attain significance. However, in alignment with the choice of C as overall preferred candidate, it is notable that, although six participants ultimately favored BERT A (M = 2.23, SD = 1.04), when it came to satisfaction, only one gave a more positive response to A than C.

In respect of perceived trust, the tests revealed a statistically significant increase with BERT C (M = 2.66, SD = 1.42), when compared with BERT B (M = 2, SD = 1.04); Z = 2.658, p = 0.008. There was also a significant difference between BERTs A (M = 2.57, SD = 1.20) and B (Z = 2.658, p = 0.002), with BERT A receiving the higher ranking. However, no significance was found in trust ratings between A and C.

The amount of frustration experienced by participants was also significantly more in the B condition (M = 2.80, SD = 1.03) than with BERT C (M = 2.23, SD = 1.22); Z = -2.546, p = 0.11). However, there was no significance found between conditions A (M = 2.52, SD = 1.20) and C. Post hoc test results are summarized in Table 2.

TABLE II. WILCOXON SIGNED RANK TESTS ON POST-EXPERIMENT QUESTIONNAIRE DATA

| Condition | Pairs | | |
|---|---|---|---|
| | **BERT A/B** | **BERT B/C** | **BERT A/C** |
| Temp. dem. | n.s. | p < .023 | p < .005* |
| Satisfaction | p < .025 | p < .005* | p < .020 |
| Trust | p < .002* | p < .008* | n.s. |
| Frustration | n.s. | p < .011* | n.s. |

* (elevated) significant findings, using Bonferroni

C. *Behavioral Data*

Behavioral and interview data were broken down into themes that emerged during the interactions and this framed the analysis.

*1) Initial impressions and engagement*

In total, five of the 21 participants attempted to talk to BERT's A and B, at least initially, even though they had been told these robots were not able to respond. Two of these participants were particularly voluble, saying things like: *"You want me to lift [the egg] out? Wow that's very impressive… Shall I take it from you this way? Very kind. Thank you so much."*

The robot was monitored much more frequently initially, with an average of three glances at the interface during the first egg handover, indicating that participants made their evaluations early in the interaction. Increased monitoring intensity was also apparent in participants moving their heads more initially to follow the robot's movements. This could be interpreted as a sign of increased presence in the interaction but is also a way to facilitate control [3].

When BERTs B and C dropped the third egg, at least five of the participants attempted to help the robot and some tried hard to prevent the eggs from falling, even succeeding in 'rescuing' the dropped egg on two occasions. These participants all went on to choose BERT C. They provided some of the highest satisfaction scores, implying a richer interaction experience. The absence of an explicit pressure to perform, as this was a home (as opposed to work) setting, may also have influenced their experience.

*2) Reactions to the "different" handover*

After BERTs B and C dropped an egg, they were programmed to try the handover again, but differently, in an attempt to rectify the situation. BERT C was able to warn participants of this, saying: *"Let's try something different."* However, almost all the participants were surprised by BERT B's 'different handover,' necessitating a rapid realignment of the way they received the egg. In both conditions B and C, they were equally in the dark about exactly how the robot would deliver it. This was manifest in tense posture and often in startled facial expressions. But their appraisals of the different handover method, in interview, varied widely. Four participants were impressed that the robot seemed to make an attempt to rectify the issue. The maneuver was also given as a reason for choosing BERT C above the rest. However, there was wide disparity in attitudes to the 'different' handover and the efficacy of the delivery in particular.

*3) Speech recognition*

More than a third of the participants were visibly irritated by problems with speech recognition which occurred in at least nine of the 21 runs with BERT C. One participant had to say yes four times before the robot released the egg and postural analysis indicated distress. However, along with the others that chose BERT C, he still welcomed voice interaction, explaining: *"With C, there was a perceived opportunity to correct something as you can answer yes or no, even one-way communication makes a big difference."*

*4) Reactions to BERT C's demonstration of regret*[3]

The contrast in nearly all the participants' reactions between BERT B's 'mistake,' with no apology or facial distress portrayed (Fig. 3, left), and BERT C's *"I'm sorry"* and exaggerated look of sadness (Fig. 3, right) was marked. A visible reaction, could be witnessed among at least six of the participants, as can be seen in Fig. 4. Three even appeared to mirror the robot's expression. In interview, its effect was also remarked upon. For example: *"You couldn't help feeling sympathy when it dropped the egg. You see the face and just go 'awww!'"*

---
[3] All images used with participant consent.

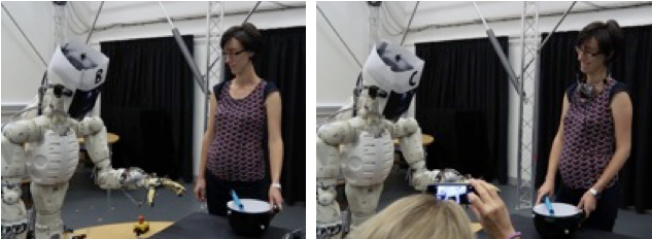

*Figure 3: The contrast in a participant's reaction to the dropped egg in conditions B (left) and C (right).*

*5) Challenging behavior*

At the end of the interaction with BERT C, when the robot asked participants whether it had performed well and if it had got the job, it made a number of them visibly uncomfortable. In interview, one explained: *"It felt appropriate to say no, but I felt really bad saying it. When the face was really sad, I felt even worse. I felt bad because the robot was trying to do its job."* His perception that BERT C's face was sad when it didn't get the job was interesting as this wasn't, in fact, the case; the robot wasn't programmed to show any form of reaction here. However, it was clear that its expressiveness made a distinct impression on him: *"In later tasks, I think I would be even more forgiving because it had expressed those emotions... The expressions did affect me, it was surprising. Once it's expressed emotion, it triggers something."*

In total, five participants reported that the feeling of being put on the spot was exacerbated by the fact that, if they told the robot it didn't perform well and didn't get the job, they couldn't qualify their answer. It was clear that subjects felt the need to say something more than a straight *"no."*

One participant appeared to be particularly reluctant to disappoint BERT C by not giving it the job. Her first answer was a *"maybe,"* and when the system didn't recognize this, she said *"yes"* but ultimately chose BERT A as her preferred job candidate in the post-experiment questionnaire. Asked about her reaction in the interview, she replied: *"Freaky! I couldn't say anything other than yes or no. [I] felt uncomfortable."* Another participant felt strongly enough to write *"emotional blackmail"* on his notepad, with the second word underlined.

*6) Temporal factors*

Participants reported that the time the task took with BERT C seemed shorter, even though it was actually approximately two minutes longer than with the other robots. In interview, three of the participants said that they thought A *"seemed slower"* or *"quite slow,"* compared to C or even B. However, some did accurately judge the longer length of the task with C. One subject who chose A as his preferred candidate said he would have chosen C, but it was *"too slow."* Another was under the impression that, compared to A, B *"was faster, more responsive and did improvise."* In fact, the interactions with A and B were approximately the same length, despite B dropping an egg.

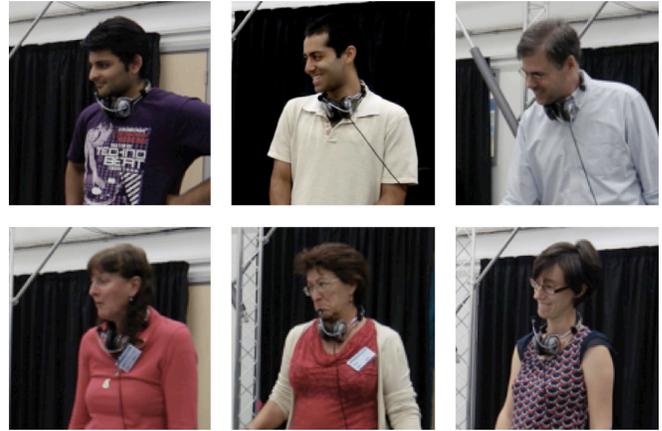

*Figure 4: Reactions to BERT C's apology and regretful expression.*

## V. DISCUSSION

This section examines the effect of the results on our Hypotheses and the implications for the design of future human-robot teams. Analysis produced a number of new avenues of enquiry, and these are also summarized together with limitations.

### A. Unexpected behavior and/or error does not need to negatively affect trust if allied with transparency.

Hypothesis 1 was that an unforeseen occurrence will cause a robot to appear less trustworthy than a more reliable one, even if attempts are made to mitigate the mistake. Supporting evidence was found for this in that BERT A was far more popular than BERT B, which was not chosen by any of the participants as their preferred candidate in the task.

Although there was no significance in the self-reported trust ratings between BERTs A and C, C was rated significantly higher than B. Its apology, expression of regret and attempt to express reparation served to forewarn participants. The fact that it appeared to try to rectify the situation and had noticed its mistake was appreciated. These actions may even have resulted in some participants increasing the level of cognitive ability they had previously ascribed to the robot, particularly if the new form of handover was seen as a successful measure [22]. This increase in perceived ability was evident in factors such as one respondent's perception that the robot seemed sad when it was not given the job, when there was no reason to surmise this.

Lower frustration levels for C as opposed to B, with correspondingly higher satisfaction and trust scores, and a higher overall rating, support Hypothesis 2: that increased transparency and feedback, manifest in communication and facial expression, can significantly mitigate dissatisfaction in the event of an unforeseen occurrence. Our results also give further credence to [28], suggesting that the effects of good automation etiquette can be powerful enough to overcome low reliability and that transparency and control could be more important to users than increased autonomy [16].

### B. The majority of users welcome, even limited, spoken communication with an assistive robot

As BERT C was rated higher overall than BERT A and achieved higher satisfaction ratings, we were also able to maintain Hypothesis 3: Given the choice between enhanced efficiency and reliability and a personable, communicative interface, most people will choose the later.

Participants' desire to talk, even to the uncommunicative robots, as well as their greater levels of engagement in the interaction and higher ratings given to BERT C, all serve to underscore a preference for an interaction that involves speech, in accordance with [15, 30]. It would imply that they are prepared to sacrifice quite a large degree of efficiency for transparency and feedback, whether the situation is a domestic or workplace one, and are content to repeat themselves if this will aid their understanding of what is happening. The problems experienced with speech also underline that increased multifunctionality comes at a cost, but with the inevitable advances in speech recognition, condition C would be likely to gain even higher preference.

Notably, the majority of participants did not mind - or even appear to notice - that the interaction with BERT C was 50 per cent longer than with A or B. Only one participant said they would have chosen C if it had not taken so long and one even thought BERT A took longer to complete the task. This would indicate that the participants' degree of involvement with BERT C was significantly higher than with the other robots.

### C. Human-like attributes can effectively smooth a difficult interaction

In condition C, communication was supplemented by further feedback in the form of the robot's expression. This had a visible effect on participants, giving further weight to studies on the importance of nonverbal communication. Its facial expression increased the robot's believability and immediately alerted participants that it "knew" it had made a mistake during a key phase in the cooperative interaction. Such a situation can be crucial to the development of trust [30]. Expressive features would thus seem a natural way for a collaborative humanoid robot to convey its intentions, particularly in a situation where it can no longer ensure a good outcome of its actions.

Some degree of mirroring of BERT C's expression was observed in at least three of our participants which could even indicate emotional contagion, the notion that when people unconsciously mimic their companions' expressions of emotion, they come to feel reflections of these [12]. This is a powerful force, as emotional contagion can serve to increase understanding and provide a form of glue for personal relationships, as well as alleviating frustration and stress [29]. However, without a more precise, electromyographical, study it is difficult to distinguish between primitive empathy, emotional contagion, and the more cognitive, sophisticated and 'social beneficial' processes of empathy and sympathy [9].

### D. Emotional and/or challenging behavior by robots must be used with care

In our experiment, participants were very reluctant to deny BERT C the job outright. At least nine appeared to be disarmed by the robot's question and at a loss as to how best to answer, attempting to modify their responses (even though this was impossible). Such a question could certainly be perceived as a "face threatening" act, as defined by [6] and could therefore have compromised their responses, although not our results: in the follow-up questionnaire, participants could answer at liberty and one even changed her answer, awarding a different robot the job. The question was a breach of established protocol and demonstrated a lack of understanding about the interview process on the part of the robot. It made no allowance for the fact that participants may not yet have "interviewed" all the robots and was there in order to assess their reactions and follow up responses. The results suggest that, having seen the robot display human-like emotion when the egg dropped, some participants were now pre-conditioned to expect a similar reaction and therefore hesitated to say no. Having developed a degree of empathy towards it, they were thus mindful that it could display further human-like distress.

### E. Limitations and future work

The short term, experimental, nature of our research precludes an exhaustive causal explanation for the observed effects, particularly BERT C's provocative question at the end of the interaction. There are very few HRI studies on robot behavior which appears to "cross the line," making this an area ripe for careful investigation. How should this type of behavior be defined? What metric should be used? Will robots end up with different personalities, just like people, depending on who designed them?

Equally interesting is our observation that people will lie to robots. It is unclear, at present, whether the intention was to avoid distress to themselves, the machine or to both, but it has potentially serious implications: for example, in the design of a robot that checks whether an elderly person has taken their medicine. In this circumstance, is it better to design robots that appear humanlike or not?

However, implementing 'personality' in collaborative robots, the design of a "new human-machine cooperative system" [36], is problematic while adequate speech recognition systems are still in their relative infancy. Our results indicate that combining speech with additional measures, such as expressive facial features, merits exploration and further research is needed to focus on specific design characteristics.

Also worthy of follow up is the impact of performance pressure in a more demanding situation, such as collaborative manufacturing. Adaptive robot behavior has been shown to be beneficial in a search and rescue scenario, for example [34]. However, contrasting reliability with expressiveness is challenging while the design of a flawless system remains out of reach.

## VI. CONCLUSION

Anyone involved in the design of present day collaborative robots is aware of the inevitability of malfunction, hence the need to explore the efficacy of compensatory measures in HRI. Building on research which used simulated situations to evaluate the effectiveness of an expressive system [27] and that which focused on verbal instruction to explore the impact of erroneous behavior [32], our work exposed users to direct physical interaction with a robot assistant in a safe environment. We found that an expressive robot was preferred over a more efficient one, despite a trade off in time taken to do the task. Participants' scores also indicated that they felt rushed with the more efficient robot, which would suggest that they placed less value on task performance and more on transparency, control and feedback, despite at least nine instances of speech recognition failure. Satisfaction was significantly increased in the communicative condition, and participants were appreciative of behavior, which they interpreted as responsive. Humanlike attributes, such as regret, were shown to be powerful tools in negating dissatisfaction but also to have a negative effect if the behavior is deemed to cross a line. Our study combined the use of self-report to evaluate participants' perceptions of the interaction with detailed analysis of key incidents where it was problematic or appeared most natural and believable. It complements prior trust-related HRI research that aims to aid the design of more reliable, acceptable and trustworthy robot companions and offers evidence that judiciously incorporating human-like attributes can significantly mitigate dissatisfaction arising from unexpected or erroneous behavior.

ACKNOWLEDGMENT

The authors are grateful for the input of BRL's Alexander Lenz, who designed and built much of BERT2, Alan Broun, who provided critical support at a difficult juncture, and Evgeni Magid, who was responsible for programming the robot for this experiment.

REFERENCES

[1] N. Bagheri, and G.A. Jamieson, G. A. "Considering subjective trust and monitoring behavior in assessing automation-induced 'complacency'", *Human performance, situation awareness, and automation: Current research and trends*, 2004, 54-59.
[2] D. Bazo, R. Vaidyanathan, A. Lentz, and C. Melhuish. "Design and testing of a hybrid expressive face for a humanoid robot", in IEEE *Int. Conf. on Intelligent Robots and Systems, 2010, pp. 5317-5322.
[3] N. Bianchi-Berthouze, "Understanding the role of body movement in player engagement", *HCI*, vol. *28*, n.1, 2013, pp. 40-75.
[4] M. Biswas, and J. C. Murray. "Towards an imperfect robot for long-term companionship: case studies using cognitive biases", in IEEE *Int. Conference on Intelligent Robots and Systems* 2015, pp. 5978-5983.
[5] C. Breazeal, C. D. Kidd, A. L. Thomaz, G. Hoffman, and M. Berlin, "Effects of nonverbal communication on efficiency and robustness in human-robot teamwork", in IEEE *International Conference on Intelligent Robots and Systems,* 2005, pp. 708-713.
[6] P. Brown, and S. C. Levinson, *Politeness: Some universals in language usage*, vol. 4. 1987, Cambridge University Press.
[7] A. Cuddy, *Presence: Bringing Your Boldest Self to Your Biggest Challenges*. 2015, Hachette UK.
[8] K. Eder, C. Harper, and U. Leonards, "Towards the Safety of Human-in-the-Loop Robotics: Challenges and Opportunities for Safety Assurance of Robotic Co-Workers", *in The 23rd IEEE International Symposium on Robot and Human Interactive Communication*, Edinburgh, 2014, pp. 660-665.
[9] N. Eisenberg, and J. Strayer, (Eds.). *Empathy and its development*. CUP Archive, 1990.
[10] T. Fong, I. Nourbakhsh, and K. Dautenhahn, "A survey of socially interactive robots", *Robotics and autonomous systems*, vol. *42*, n. 3, 2003, pp. 143-166.
[11] A. Hamacher, *Believing in BERT: Making good on bad robot behaviour* (Master's thesis, UCL, London, UK), 2015. Retrieved from: http://preview.tinyurl.com/zjphu8e
[12] E. Hatfield, J. T. Cacioppo, and R. L. Rapson, Emotional contagion. New York: Cambridge University Press,1994.
[13] C. Heath, J. Hindmarsh, and P. Luff, *Video in qualitative research*. Sage, 2010.
[14] S. Hutson, S.L. Lim, P.J. Bentley, N. Bianchi-Berthouze and A. Bowling, "Investigating the suitability of social robots for the wellbeing of the elderly", in ACII'1, pp. 578-587
[15] Y. Iwamura, M. Shiomi, T. Kanda, H. Ishiguro, and N. Hagita, "Do elderly people prefer a conversational humanoid as a shopping assistant partner in supermarkets?", in *Proc. of the Int. Conference on Human-Robot Interaction,* 2011, pp. 449-456.
[16] M. Johnson, J. M. Bradshaw, P. J. Feltovich, C. Jonker, B. van Riemsdijk, and M. Sierhuis, "Autonomy and interdependence in human-agent-robot teams", *Intelli. Syst.* vol. 7, n. 2, 2012, pp 43-51.
[17] B. Jordan, and A. Henderson, "Interaction analysis: Foundations and practice", *The J. of the Learning Sciences*, vol 4, n., 1995, pp. 39-103.
[18] T. Kim, and P. Hinds, "Who should I blame? Effects of autonomy and transparency on attributions in human-robot interaction", in *Int. Symp. on Robot and Human Interactive Communication,* 2006, pp. 80-85.
[19] S. Lallée, U. Pattacini, S. Lemaignan, A. Lenz, C. Melhuish, L. Natale, and P. F. Dominey, "Towards a platform-independent cooperative human robot interaction system: III an architecture for learning and executing actions and shared plans", *IEEE Transactions on Autonomous Mental Development,* vol 4, n. 3, 2012, 239-253.
[20] J.D. Lee, and K. A. See, "Trust in automation: Designing for appropriate reliance", *Human Factors: The Journal of the Human Factors and Ergonomics Society*, vol. 46, n. 1, 2004, pp. 50-80.
[21] M. K. Lee, S. Kiesler, and J. Forlizzi, "Receptionist or information kiosk: how do people talk with a robot?", in *Proc. of the ACM Conf. on Computer supported cooperative work,* 2010, pp. 31-40.
[22] S. Lemaignan, J. Fink, and P. Dillenbourg, "The dynamics of anthropomorphism in robotics", in *Proceedings of the International Conference on Human-Robot Interaction,* 2014, pp. 226-227.
[23] A. Lenz, S. Skachek, K. Hamann, J. Steinwender, A.G. Pipe, and C. Melhuish, "The BERT2 infrastructure: An integrated system for the study of human-robot interaction", in Proc. of the *Int. Conference on Humanoid Robots (Humanoids),* 2010, pp. 346-351.
[24] C. A. Miller, "Trust in adaptive automation: the role of etiquette in tuning trust via analogic and affective methods", in *Proc. of the Int. Conference on Augmented Cognition*, 2015, pp. 22-27.
[25] B. M. Muir, and N. Moray, "Trust in automation. Part II. Experimental studies of trust and human intervention in a process control simulation", *Ergonomics*, vol. 39, n. 3, 1996, 429-460.
[26] D. A. Norman, "The 'problem' with automation: inappropriate feedback and interaction, not 'over-automation'", *Phil. Trans. Royal Soc. of London B: Biol. Sci.*, vol. *327*, n. 1241, 1990, pp. 585-593.
[27] J. Novikova, L. Watts, and T. Inamura, "Emotionally expressive robot behavior improves human-robot collaboration", in the *IEEE Int. Symp. on Robot and Human Interactive Communication,* 2015, pp. 7-12.
[28] R. Parasuraman, and C. A. Miller, "Trust and etiquette in high-criticality automated systems", *Communication of the ACM*, vol. *47*, n. 4, 2004, pp. 51-55.
[29] R. W. Picard, "Building an affective learning companion", in *Intelligent Tutoring Systems*, 2006, pp. 811-811.
[30] C. Ray, F. Mondada, and R. Siegwart, "What do people expect from robots?", in *International Conference on Intelligent Robots and Systems,* 2008*,* pp. 3816-3821.
[31] D. M. Rousseau, S. B. Sitkin, R. S. Burt, and C. Camerer, "Not so different after all: A cross-discipline view of trust. *Academy of Management Review*, vol. 23, n. 3, 1998, pp. 393-404.
[32] M. Salem, G. Lakatos, F. Amirabdollahian, and K. Dautenhahn, "Would You Trust a (Faulty) Robot?: Effects of Error, Task Type and Personality on Human-Robot Cooperation and Trust", in *Proc. of the Int. Conf. on Human-Robot Interactio*n, 2015, pp. 141-148.
[33] C. L. Sidner, C. Lee, and N. Lesh, Engagement rules for human-robot collaborative interactions. In *IEEE Int. Conference On Systems Man And Cybernetics*, vol. 4, 2003, pp. 3957-3962.
[34] M. M. Sobhani, A. G. Pipe, S. Dogramadzi and J. G. Fennell, "Towards model-based robot behaviour adaptation: Successful human-robot collaboration in tense and stressful situations", in *Iranian Conference on Electrical Engineering (ICEE),* 2015, pp. 922-927.
[35] T. Trower, Personal communication, 2014.
[36] D.D. Woods, "Decomposing automation: Apparent simplicity, real complexity", *Automation and human performance: Theory and applications*, 1996, pp. 3-17.